%% file: elk-template.tex
\documentclass{arxiv}
\usepackage{hyperref}
\hypersetup{
colorlinks=true,
urlcolor=black,
citecolor=blue}
\usepackage[all]{xy,xypic}
\usepackage{amsfonts,amssymb,amsmath,amsgen,amsopn,amsbsy,theorem,graphicx,epsfig}
\usepackage{eufrak,amscd,bezier,latexsym,mathrsfs,eurosym,enumerate}
\usepackage[utf8]{inputenc}
\usepackage{float} 
\usepackage[english]{babel}
\usepackage{cleveref,multirow}
\usepackage[dvipsnames]{xcolor}
\title{
Towards Transparent Diagnostics: Investigating Architectural Trade-offs and Explainability in Malaria Detection
}

\author{
    Suman Kunwar$^{1}$ and Avishek Dangol$^{2}$
    \\[0.4em]
    $^{1}$DWaste, Baltimore, USA
    \quad
    $^{2}$University of the West of England, Bristol, UK
}

\input{elksty.tex}

\begin{document}

\maketitle

\begin{abstract}More than 80 countries have reported malaria cases with 610 thousand deaths and are projected to increase. Identifying malaria early and accurately helps save lives and effective way to diagnose malaria is through microscopic methods that are labor intensive and require experts with special equipment. Deep learning (DL) has shown promising results in medical diagnosis. Here, we explored various DL models: ResNet18, MobileNetV2, EfficientNet-B2, VGG19 and proposed model ResNet18+TTA (ResNet18 backbone with modified classification head and test time augmentation) for detecting malaria presence using blood smears taken from the NIH Malaria dataset.
         
Our experiment shows MobileNetV2 achieved 96.85\% accuracy with smallest model size (8.49 MB) and fastest inference (1.35 ms). The ResNet18+TTA model achieved 97.96\% accuracy, 0.996 AUC with longest inference time (13.32 ms). Larger architecture outputs a larger model size with moderate accuracy. Upon further pruning, ResNet18+TTA model gained a slight improvement in accuracy and reduced inference time. GRAD-CAM, SHAP and LIME provide explainable AI (XAI) insights into model predictions, using explanation agreement and divergence to evaluate predictive reliability.

\end{abstract}
\section{Introduction}
\label{Int}
Malaria is a life threatening mosquito-borne infectious disease caused by Plasmodium parasite that are found mainly in tropical countries. It gets transmitted to humans through the bites of infected female Anopheles mosquitoes, blood transfusion and contaminated needles. Within the five plasmodium parasites, P.falciparum and P.vivax pose a greater threat than the others (P.malariae, P.ovale, and P.knowlesi). P.falciparum is the deadliest and most common in the African content, whereas P.vivax is found in countries outside sub-Saharan Africa.

In 2024, an estimated 282 million malaria cases were reported in 80 countries with 610 thousands deaths \cite{who2025}. 95\% (265 million) malaria cases and 95\% (579,000) of malaria deaths were reported from African region. Among these cases, children under 5 years of age accounted for about 75\% of all malaria deaths \cite{platon2026}. With the climate change malaria is projected to grow and can lead to additional 123 million malaria cases with 532,000 deaths under current control levels in Africa between 2024 and 2050 \cite{symons2026}. Vector control is an effective way to to control and eliminate malaria, as they reduce the exposure to infected mosquito vectors thus limiting transmission \cite{who2015}. Malaria surveillance also helps design effective health interventions and evaluate malaria control programs.

Recent technological advancements have improved the ability to diagnose different species of Plasmodium with greater sensitivity and specificity. Approaches such as Microscopic methods, serodiagnostic assays and molecular methods are used in the detection of plasmodium parasites  infecting human methods \cite{slater2022}. Morphological appearances are also used to identify mosquito species for vector surveillance \cite{dasari2024}. The use of microscopic Giemsa-stained blood smears is considered the gold standard for the quantification and identification of species \cite{parveen2025}. However, it is labor intensive, time consuming, requires experts, and is not suitable for a resource constrained environment \cite{wilson2012}.

Rapid diagnostic tests (RDTs) detect specific antigens like HRP2 and pLDH produced by malaria parasites that offer a quicker alternative but their inability to distinguish species and reduced sensitivity and specificity (low parasite densities and with non-falciparum species) limits them. Polymerase chain reaction (PCR) based detection offers high sensitivity and specificity but is costly, time consuming, and requires specialized infrastructure making them impractical for widespread deployment \cite{rougemont2004}.

These limitations underscore the need for an efficient, accurate and automated approach for diagnosis purposes. The advancement in AI has opened new avenues in medical diagnosis, offering automated analysis with great precision but also raises questions about the explainability and interpretability of the model \cite{alkhanbouli2025}.  While studies have reported strong accuracy on the NIH malaria dataset, two critical weakness exists on current deep learning studies. First, most studies are evaluated on random cell level splits that leaks same slide into training and test sets thus inflating reported accuracy \cite{hou2026}. Second, explainability metrics are largely qualitative with handful sample images quantitatively assessing explanation faithfulness, stability, or failure modes \cite{parveen2025}. These two gaps motivate the study and aims to answer the following research questions.

\begin{itemize}
    \item RQ1: Which CNN architectures provides optimal accuracy–efficiency trade-off among lightweight (MobileNetV2), mid-sized (ResNet18, EfficientNet-B2), heavy (ResNet101, VGG19), and  proposed model (ResNet18+TTA) for malaria classification under slide-disjoint evaluation?
    \item RQ2: Do explainability tools such as Grad-CAM, SHAP, and LIME  provide faithful and stable explanations for malaria detection, and how do they behave on misclassified samples?
\end{itemize}
The remainder of the paper is organized as follows: Section~\ref{Rel} discusses related work, section~\ref{Mat} methodologies and materials used in this study, section~\ref{Res} describes results and findings and section~\ref{Con} concludes the study.

\section{Related Work}
\label{Rel}
\subsection{Deep Learning for Malaria Detection}
Prior the adoption of deep learning (DL), malaria diagnosis from blood-smear images was largely based on handcrafted feature extraction such as color, texture, and morphological descriptors combined with classical classifiers such as Support Vector Machines (SVM) and Random Forests (RF) \cite{poostchi2018}. These approaches were effective under controlled conditions but the pipelines were sensitive to staining variability and also required substantial domain expertise for feature engineering thus motivating the shift toward Convolutional Neural Networks (CNNs).

CNNs have since become the most preferred approach for automated malaria diagnosis, reporting state-of-the-art performance on blood-smear image datasets commonly in NIH Malaria Dataset \cite{malariadataset}. Early work demonstrated that compact models can perform well on malaria diagnosis. For instance, a four-layer CNN with two fully connected layers reported an accuracy of 99.51\% (precision: 99.26\%, recall: 99.26\%, F1-score: 99.26\%) on NIH Malaria Dataset with 80/10/10 train-validation-test split, suggesting that standard CNN feature extractors are effective for distinguishing infected from uninfected cells \cite{ramos2025}.

Subsequent work introduced complex architectures with multi-branch feature extractors and attention mechanisms. A parallel CNN with soft attention with 2.207M parameters, achieved 99.37\% accuracy (precision: 99.38\%, recall: 99.37\%, F1-score: 99.37\%) \cite{ahamed2025}. Beyond CNNs, Gaouar et al. proposed a stacked long-short-term memory network (LSTM) with attention mechanism, achieving 99.12\% accuracy (precision: 99.59\%, recall: 98.67\%, F1-score: 99.11\%) by treating image patches as sequential inputs to capture long-range spatial dependencies \cite{gaouar2025}. Transfer learning (TL) approaches using pretrained backbones such as VGG, ResNet, MobileNet, and DenseNet have also been widely applied to this task, reporting accuracies in the range of 89\% - 95.7\% \cite{rajaraman2018}.

Despite strong performance, many CNN models rely on millions of parameters, resulting in an increased memory footprint, a higher inference latency and limited suitability for deployment in low-resource clinical settings. To address these constraints, Ali et al. explored lightweight CNN with approximately 1.2M parameters, reporting a reduced accuracy of 95.45\% on the same dataset. Their results highlight the trade-off between model compactness and predictive performance but also underscore the need for architectures that balance efficiency with robustness \cite{ali2024}.

Although lightweight CNNs reduce computation, they often sacrifice predictive performance, motivating research into compression techniques that preserve accuracy while reducing inference cost. Pruning removes redundant weights or channels from a network, enabling deployment on resource-constrained devices \cite{han2015a}. Early work on magnitude-based and filter pruning demonstrated that large CNNs can be compressed by a range of 30× to 49× with minimal accuracy loss, enabling real-time inference on mobile hardware \cite{han2015b,li2016}.

\subsection{Explainable AI in Malaria Detection}
Since CNNs operate as high-dimensional, non-linear feature extractor, their decision making process is often opaque, making it difficult to determine which image regions drive a prediction \cite{samek2017}. This ‘black-box’ nature has motivated extensive work in explainable AI (XAI) for medical imaging. Techniques such as Grad-CAM (Gradient-weighted Class Activation Mapping), LIME (Local Interpretable Model-Agnostic Explanations), and SHAP (SHapley Additive exPlanations) have been applied to malaria detection models to highlight salient regions, assess feature importance, and provide post-hoc interpretability \cite{awe2025,islam2022}. For example, Ahmed et al. applied Grad-CAM and SHAP to a CNN trained on the NIH dataset. Grad-CAM highlighted coarse activation regions centered on parasite-containing areas, providing spatial evidence of the model’s focus \cite{ahamed2025}. SHAP, in contrast, offered a finer-grained attribution by identifying specific pixel and superpixel contributions, showing that the features driving the model’s predictions corresponded to parasite-associated structures rather than irrelevant artefacts.

Similarly, Gaouar et al. used LIME to identify critical features that contribute most to individual predictions, finding that LIME emphasizes fine-grained boundary information by highlighting the cell’s contour regions, and is complemented by Grad-CAM which produce a more holistic representation by mapping the broader activation patterns distributed across the entire cell \cite{gaouar2025}.  Attai et al. investigate interpretability by integrating LIME with Large Language Models experimenting among ChatGPT, Gemini, and Perplexity and reporting that these approaches can enhance clinicians’ acceptance of AI-assisted diagnostics by increasing transparency in model reasoning. However, they also note that such methods may complicate real-time deployment due to their substantial computational requirements and dependence on stable internet connectivity \cite{attai2024}.

Building on the earlier discussion of pruning as a compression strategy, recent work has also examined its implications for interpretability. Merkle et al. investigated how pruning affects attribution quality and showed that aggressive parameter removal can distort Grad‑CAM explanations, underscoring the need for compression strategies that preserve both predictive fidelity and explanation stability \cite{merkle2025}. Ghosh et al. extended this line of inquiry to a clinical context and reported that the unpruned model, retaining 100\% of its weights, correctly identified the top three clinically relevant concepts associated with skin-lesion malignancy. In contrast, when the model was pruned to 23\% of its original weights, it instead surfaced a clinically irrelevant concept as one of the key indicators of malignancy \cite{ghosh2023}.
 
\section{Material and Methods}
\label{Mat}
\subsection{Dataset and Preprocessing}
\begin{figure}[htbp]
\begin{center}
\includegraphics[ width=0.4\textwidth]{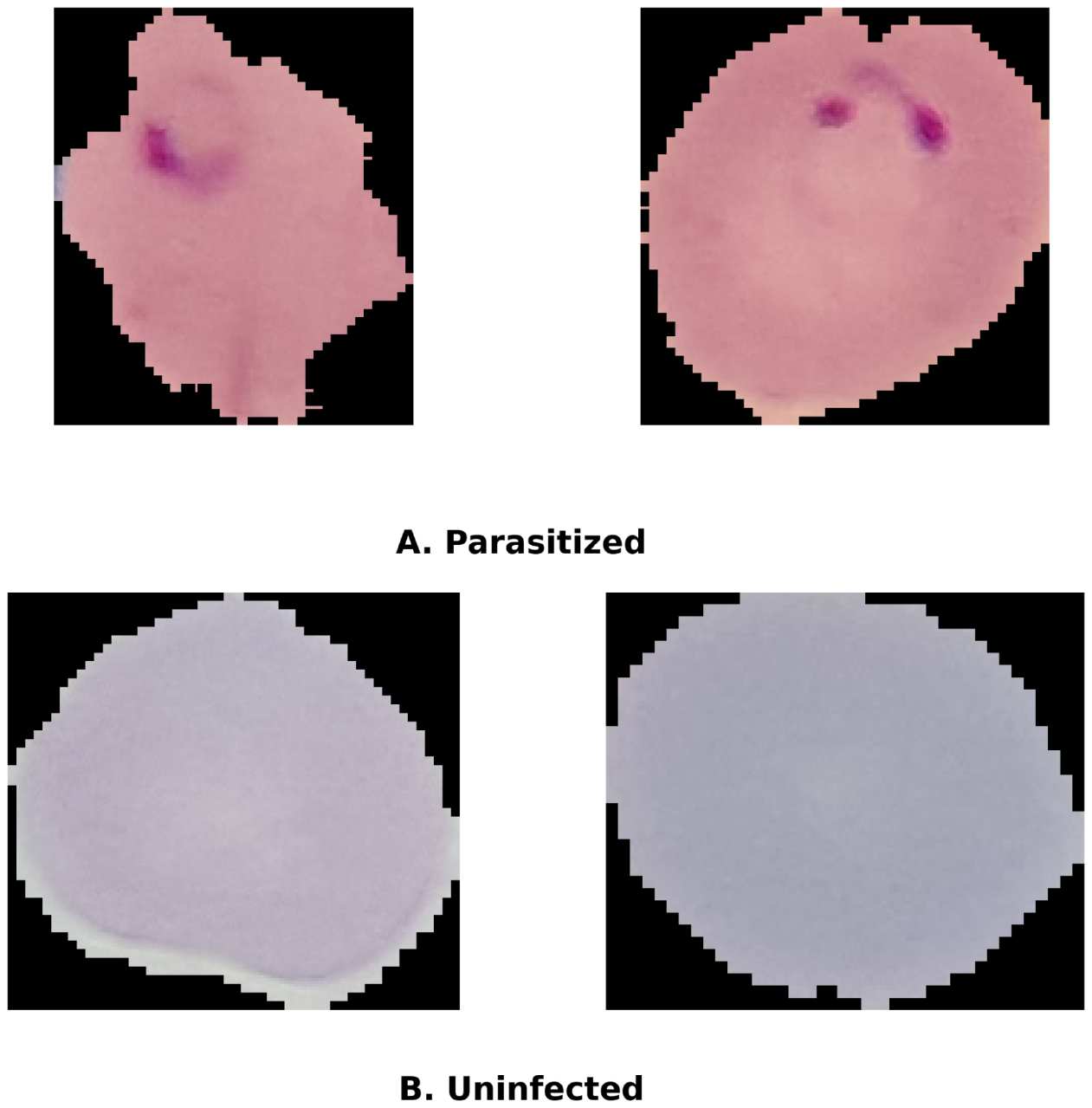}
\caption{Sample images of Parasitized and Uninfected cells}
\label{fig:cell_samples}
\end{center}
\end{figure}

This research uses NIH malaria dataset consisting of 27,558 malaria cell images equally spitted into infected (Parasitized) and uninfected across 200 slides and all slides contains both classes. These images were created using giemsa-stained thin blood smear slides obtained from 150 infected and 50 healthy patients and were annotated by experts. An example of sample images from the dataset is shown in Figure \ref{fig:cell_samples}, where we can see that the image sizes are inconsistent.

\begin{figure}[H]
\begin{center}
\includegraphics[width=0.9\textwidth]{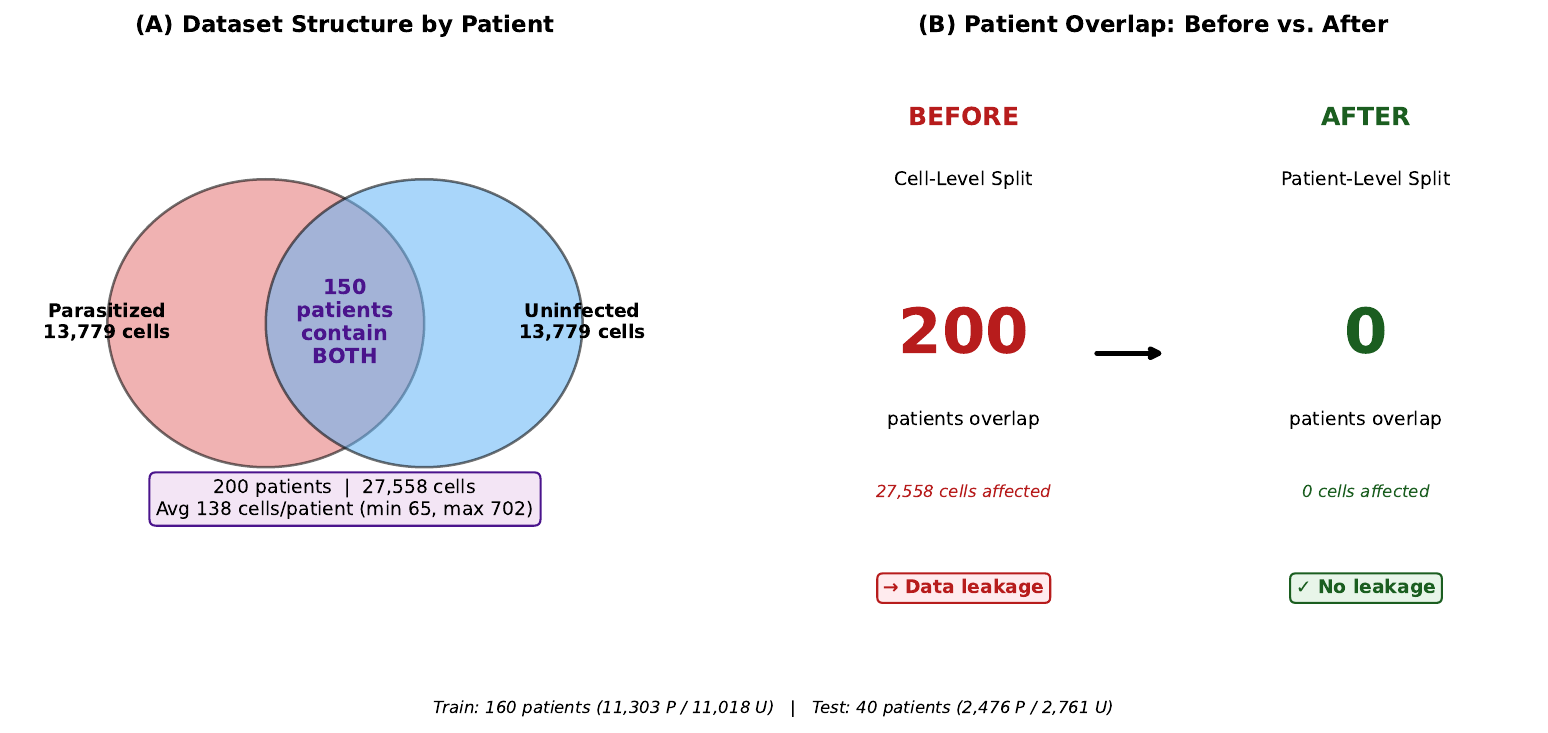}
\caption{Dataset structure and leakage verification: A) Dataset Structure by Patient; B) Patient Overlap: Before vs. After}
\label{fig:patient_structure_and_leakage}
\end{center}
\end{figure}

The infected collection contains parasite images with various morphological appearances representing stages of malaria parasite, whereas the uninfected collection contains images with artifacts such as dust particles and uneven staining representing  data diversity and complexity in the data. Images are grouped based on slide identifier extracted from filenames to prevent data leaks in such a way that same patient do not appear in multiple splits as shown in Figure \ref{fig:patient_structure_and_leakage}. This is equivalent to slide-disjoint, since each patient corresponds to one slide.

Using cell split, all  200 patients appear in both training and test sets 27,558 causing data leakage whereas using patient-level grouping, zero slide overlaps. The final split based on 80/20 training and test sets contains 160 training slides and 40 test slides. Augmentation techniques such as flipping, rotation and color jitter are applied  for data enrichment. To ease the reproducibility of experiments, the random seed was fixed (seed = 42). 

\subsection{Model Training and Evaluation}
\begin{figure}[H]
  \centering
  \includegraphics[width=0.4\textwidth]{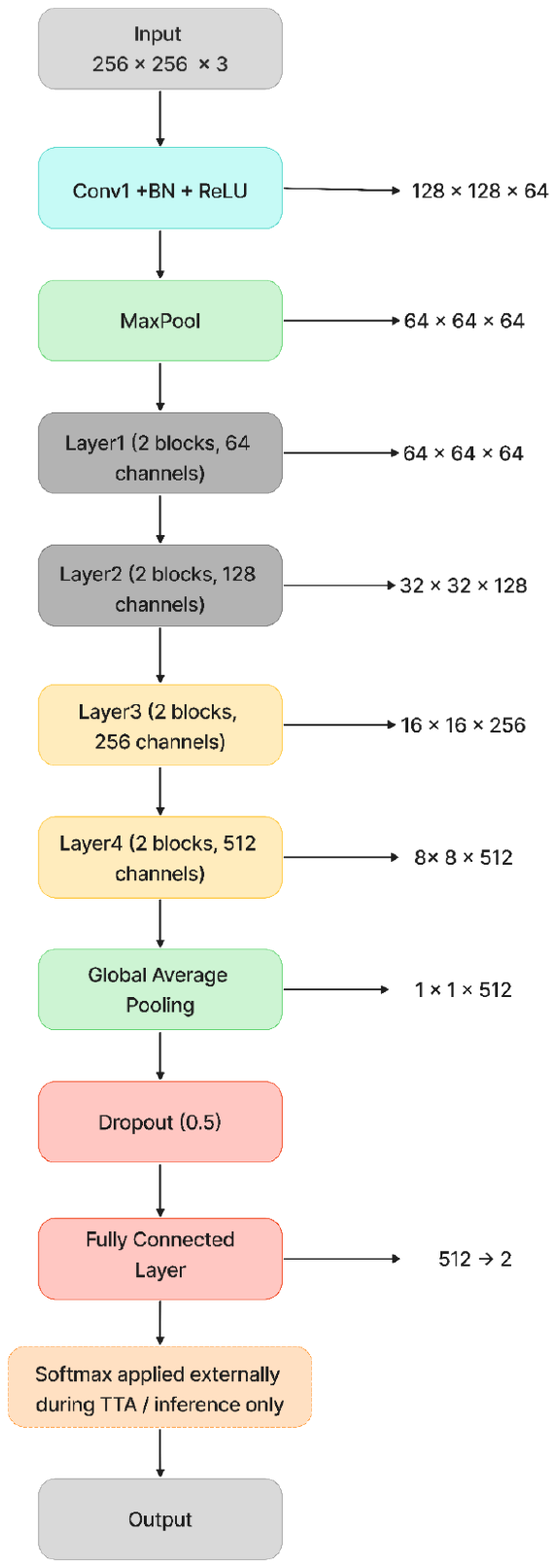} % Adjust 0.7 to 0.5–0.8 as needed
  \caption{ResNet18+TTA model architecture}
  \label{fig:proposed_model}
\end{figure}

Deep learning models learn patterns through their layers and work well on large datasets. Models with complex architecture often consume a lot of computational resources, which are often fine tuned to make it efficient. Here, we proposed a modified ResNet18 model that uses ResNet18 backbone with a custom classification header and a as shown in Figure \ref{fig:proposed_model}. We used the ResNet18 variant as it is the smallest standard ResNet variant and an established NIH baseline.

The custom classifier head consists of a Linear (512, 2) layer to output law logits for binary classification. All convolution blocks (layer 1 to layer 4) are fully trainable, and the activation function is applied during the inference stage. The Test Time Augmentation (TTA) is applied to access whether the inference-time strategy can improve robustness. The proposed model (ResNet18+TTA ) is evaluated with other state of the art malaria classification  models: MobileNetV2, EfficientNet-B2, ResNet101, VGG19 and ResNet18, and ablation study of ResNet18 is also conducted. The training was performed on Kaggle using Tesla x T2 GPU, for 15 epochs. Standard metrics such as Accuracy, Precision, Recall, F1-score, AUC, parameter count, model size, and inference time are used for benchmarking.

\subsection{Explainability Measures}
The explainability of the best performing AI model is also explored using the following approaches.
\begin{itemize}
    \item Gradient-weight Class Activation Mapping (Grad-CAM): It is an extended version of CAM used to examine the spatial region that has strong influence towards model prediction. It is valuable in medical diagnosis as it validates that the model is learning important clinically relevant features \cite{selvaraju2017}. 
    
    % It is computed as

    % \begin{equation}
    % \alpha_k^c = \frac{1}{Z} \sum_{i} \sum_{j} \frac{\partial y^c}{\partial A_{i,j}^k}
    % \end{equation}
    
    % \begin{equation}
    %     L_{\text{Grad-CAM}}^c = \text{ReLU} \left( \sum_{k} \alpha_k^c A^k \right)
    % \end{equation}
    
    % where:
    % \begin{itemize}
    %     \item[] $c$ = predicted class of interest
    %     \item[] $\alpha_k^c$ = importance weight of feature map $k$ for class $c$
    %     \item[] $A^k$ = activation map of channel $k$
    %     \item[] $y^c$ = score for class $c$ before softmax layer
    %     \item[] $Z$ = global average pooling factor (width $\times$ height of feature map)
    %     \item[] $A_{i,j}^k$ = pixel intensity at position $(i, j)$ in feature map $A^k$
    % \end{itemize}

    \item Local Interpretable Model-Agnostic Explanations (LIME): It operates on model agnostic interpretability principle and provides an explanation of predictions by evaluating the impact of localized perturbations on the output. This method helps to validate that the model decisions are based on the meaningful biological characters rather than background noise or irrelevant artifacts \cite{ribeiro2016}. 
    % It is defined as

    % \begin{equation}
    % \text{explanation}(x) = \arg\min_{g \in G} \mathcal{L}(f, g, \pi_x) + \Omega(g)
    % \end{equation}
    % where:
    % \begin{itemize}
    %     \item[] $f$ = original model being explained
    %     \item[] $x$ = specific instance being explained
    %     \item[] $G$ = a class of interpretable models (e.g., linear models, decision trees)
    %     \item[] $g \in G$ = candidate interpretable model
    %     \item[] $\pi_x$ = proximity measure defining how large the neighborhood is around $x$ that we're explaining
    %     \item[] $\mathcal{L}(f, g, \pi_x)$ = fidelity function measuring how unfaithful $g$ is in approximating $f$ within the neighborhood defined by $\pi_x$
    %     \item[] $\Omega(g)$ = complexity penalty for the interpretable model $g$
    % \end{itemize}

    \item SHapely Additive exPlanations (SHAP): It provides fine grained interpretation of models prediction as it quantifies the contribution of individual input features to the final model's output using Shapley values derived from Game theory. Attribution score is assigned to each pixel to identify the regions that positively or negatively influence the predicted class \cite{lundberg2017}. 
    % It is computed as 

    % \begin{equation}
    % \phi_i = \sum_{S \subseteq F \setminus \{i\}} \frac{|S|!(|F| - |S| - 1)!}{|F|!} \left[ f_{S \cup \{i\}}(x_{S \cup \{i\}}) - f_S(x_S) \right]
    % \end{equation}
    % where:
    % \begin{itemize}
    %     \item[] $F$ = set of all features
    %     \item[] $S$ = a subset (coalition) of features not including $i$
    %     \item[] $x_S$ = input where only features in $S$ are present
    %     \item[] $f_{S \cup \{i\}}(x_{S \cup \{i\}}) - f_S(x_S)$ = marginal contribution of feature $i$ to coalition $S$
    %     \item[] The fraction $\frac{|S|!(|F| - |S| - 1)!}{|F|!}$ = a weighting term representing all possible orderings in which feature $i$ could be added to the coalition $S$
    % \end{itemize}

\end{itemize}

% Compared to Grad-CAM and LIME, SHAP provides more detailed and quantitative information on the pixel level.

\section{Results and Discussion}
\label{Res}
From the experiments shown in Table \ref{tab:benchmark}, lightweight architectures such as MobileNetV2 with the smallest model size 8.49 MB achieved fastest inference time of 1.35 ms/image. Whereas the larger architecture such as ResNet101 and VGG19 not only have the larger model size but the inference time is also longer. VGG19 achieves highest Recall (0.9898) but also the largest model size 532.45 MB. 
The EfficientNet-B2 with moderate parameters (7.70 Million), achieved 97\% accuracy, with medium model size (29.39 MB) with 2.13 ms inference speed.
ResNet18 provided comparable accuracy (96.92 \%), F1-score (0.966) at lower latency (1.42 ms) and is therefore selected as a backbone for the ablation study in Table \ref{tab:ablation} where the TTA effect is analysed.

\begin{table}[H]
\caption{Performance comparison of different backbones.}
\label{tab:benchmark}
\centering
\small
\setlength{\tabcolsep}{4pt}
\renewcommand{\arraystretch}{1.05}
\begin{tabular}{lcccccccc}
\hline
\textbf{Model} &
\textbf{Acc.} &
\textbf{Prec.} &
\textbf{Rec.} &
\textbf{F1} &
\textbf{AUC} &
\textbf{Param.} &
\textbf{Size} &
\textbf{Time} \\
 &
\textbf{(\%)} &
 &
 &
 &
 &
\textbf{(M)} &
\textbf{(MB)} &
\textbf{(ms)} \\
\hline
ResNet18        & 96.92 & 0.949 & 0.984 & 0.966 & 0.993 & 11.18  & 42.64  & 1.42 \\
MobileNetV2     & 96.85 & 0.951 & 0.980 & 0.965 & 0.990 & 2.23   & 8.49   & 1.35 \\
EfficientNet-B2 & \textbf{97.00} & \textbf{0.953} & 0.982 & \textbf{0.967} & 0.991 & 7.70 & 29.39 & 2.13 \\
ResNet101       & 96.75 & 0.951 & 0.977 & 0.964 & 0.990 & 42.50  & 162.14 & 5.79 \\
VGG19           & 96.69 & 0.939 & \textbf{0.990} & 0.964 & \textbf{0.993} & 139.58 & 532.45 & 6.65 \\
\hline
\end{tabular}

\vspace{2pt}
\begin{minipage}{\linewidth}
\footnotesize
\textit{Note:} All models were trained with AdamW (LR $5\times10^{-5}$,
weight decay $1\times10^{-4}$), ReduceLROnPlateau scheduling,
cross-entropy loss with label smoothing 0.1, $224\times224$ inputs,
and no MixUp or gradient clipping. Inference time is per image.
\end{minipage}
\end{table}

Table \ref{tab:ablation} shows the ablation study on ResNet18, where TTA combined with ResNet18 achieved highest accuracy (97.96 \%), F1-Score(0.9806) and AUC (0.9966), a 0.02 pp gain over the baseline. Dropout reduces the accuracy on both cases (standalone and combined with TTA). All combination fell with ~0.08 pp, indicating small differences. The dominant effect is that, TTA increase the inference cost.

\begin{table}[H]
\caption{Effect of Dropout and TTA on ResNet18.}
\label{tab:ablation}
\centering
\small
\begin{tabular}{lcccccc}
\hline
\textbf{Configuration} &
\textbf{Acc. (\%)} & \textbf{Prec.} & \textbf{Rec.} &
\textbf{F1} & \textbf{AUC} & \textbf{Time (ms)} \\
\hline
(1) ResNet18 baseline  & 97.94 & \textbf{0.9808} & 0.9801 & 0.9804 & 0.9965 & \textbf{1.67} \\
(2) + Dropout(0.5)     & 97.88 & 0.9783 & 0.9815 & 0.9799 & 0.9962 & 1.70 \\
(3) + TTA              & \textbf{97.96} & 0.9798 & 0.9815 & \textbf{0.9806} & \textbf{0.9966} & 13.32 \\
(4) + Dropout + TTA    & 97.94 & 0.9787 & \textbf{0.9823} & 0.9805 & 0.9965 & 13.36 \\
\hline
\end{tabular}

\vspace{2pt}
\begin{minipage}{\linewidth}
\footnotesize
\textit{Note:} All configurations use the ResNet18 backbone and were trained with AdamW
(LR $1\times10^{-4}$, weight decay $1\times10^{-4}$),
CosineAnnealingWarmRestarts scheduling, cross-entropy loss with label
smoothing 0.1, MixUp ($\alpha=0.2$), gradient clipping (norm 1.0),
$256\times256$ inputs, and early stopping (patience 5).
\end{minipage}
\end{table}

\begin{figure}[H]
\begin{center}
\includegraphics[width=0.8\textwidth, trim=0cm 0cm 0cm 0cm, clip]{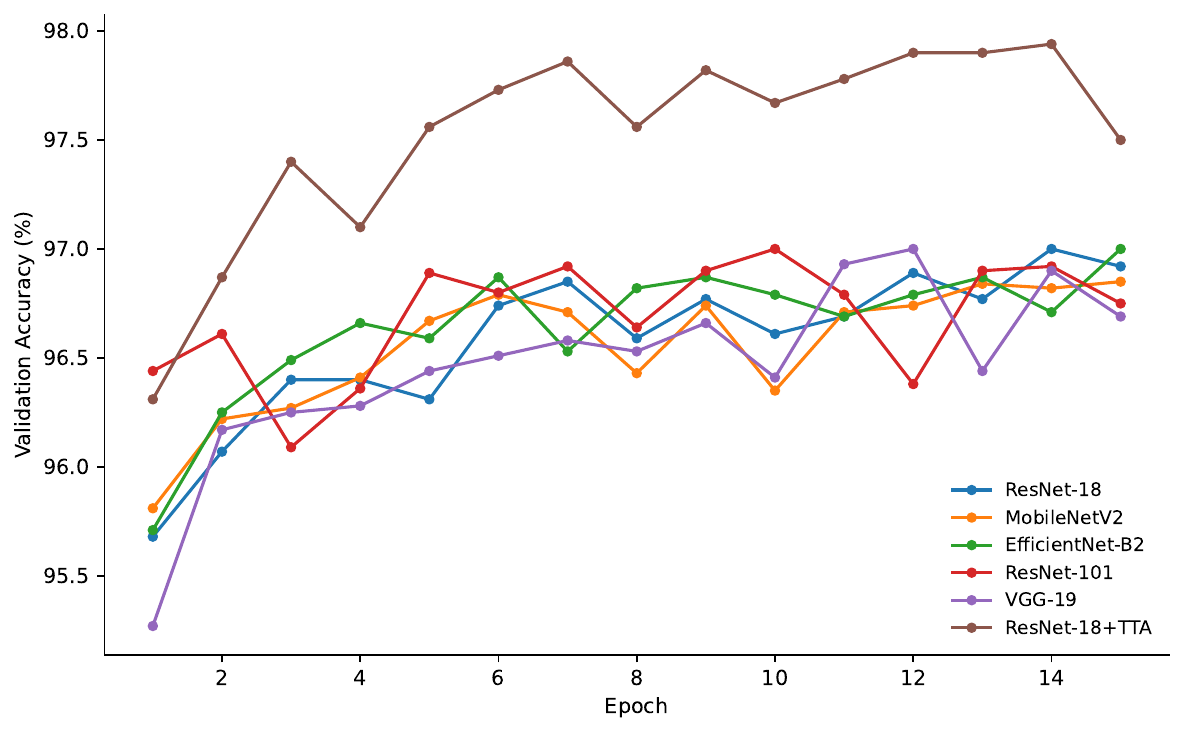}
\caption{Validation accuracy during model training}
\label{fig:validation_accuracy}
\end{center}
\end{figure}

\begin{figure}[H]
\centering
\includegraphics[width=0.8\textwidth, trim=0cm 0cm 0cm 0cm, clip]{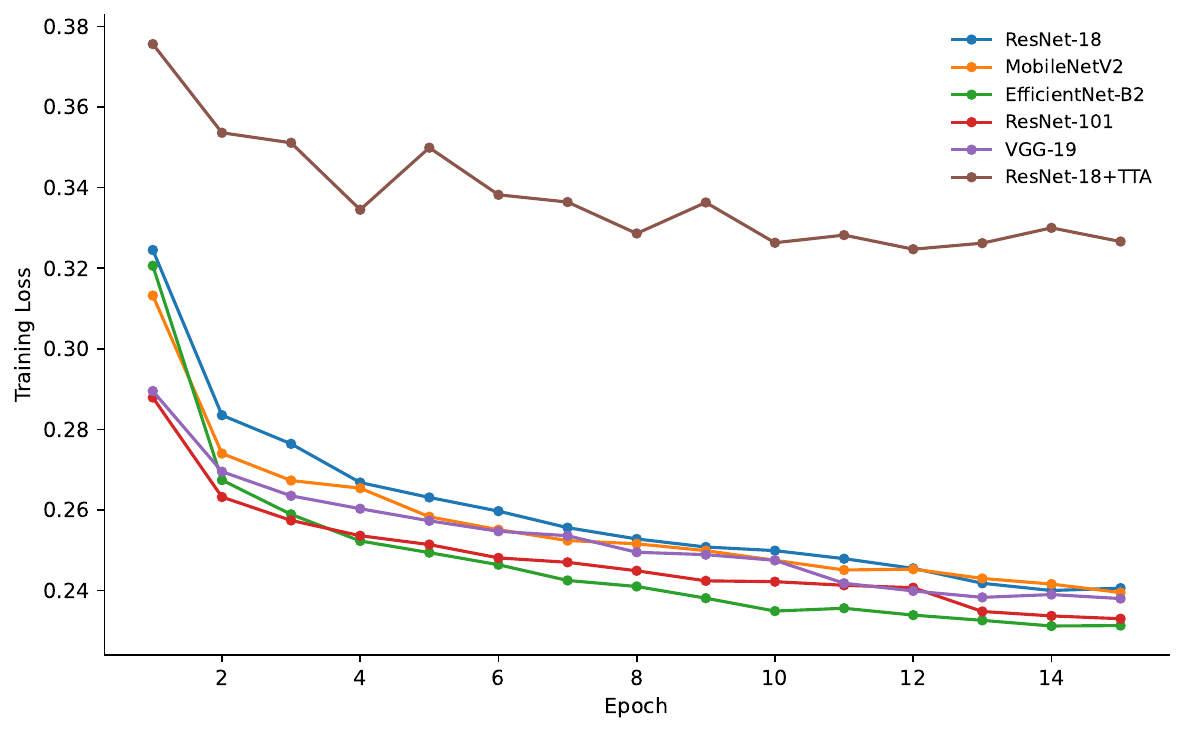}
\caption{Training loss during model training}
\label{fig:training_loss}
\end{figure}

The validation accuracy and training loss during model training are shown in Figure \ref{fig:validation_accuracy} and Figure \ref{fig:training_loss} respectively, where we can see that the ResNet18+TTA model performs well from the beginning despite the higher training loss, indicating strong generalization and resistance to overfitting.

\begin{figure}[H]
\centering
\includegraphics[width=0.5\textwidth]{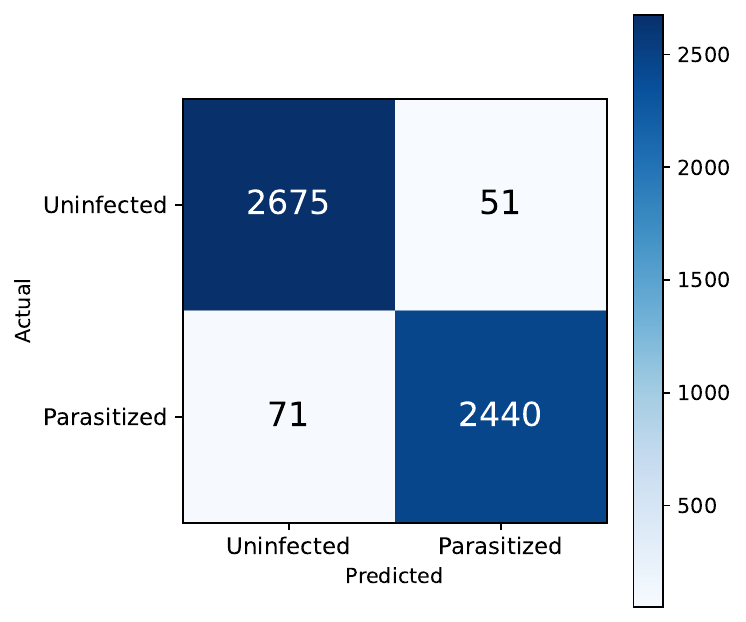}
\caption{Confusion Matrix of the ResNet18+TTA model}
\label{fig:confusion_matrix}
\end{figure}

\begin{figure}[H]
\centering
\includegraphics[width=0.5\textwidth]{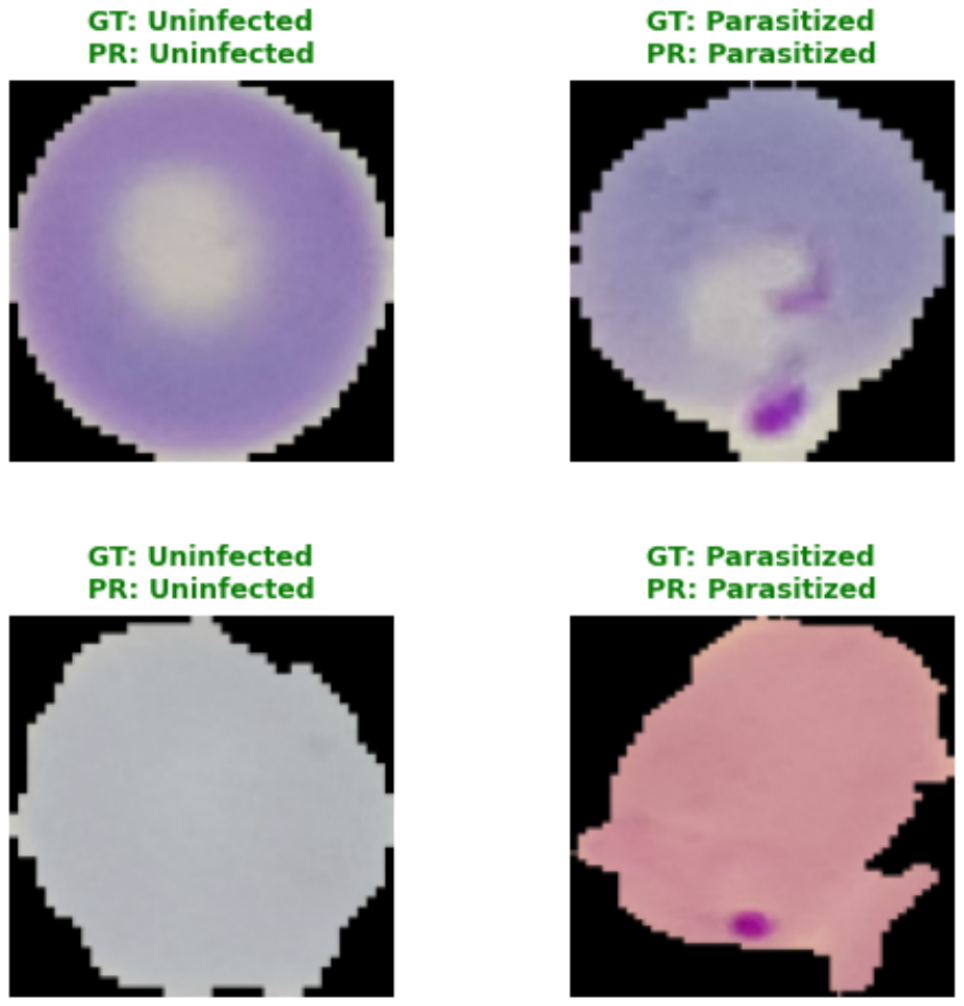}
\caption{Prediction on the sample validation images using ResNet18+TTA model}
\label{fig:sample_prediction}
\end{figure}

The confusion matrix of the actual and predicted value is shown in Figure \ref{fig:confusion_matrix}, where the ResNet18+TTA model predicted 2,675 Uninfected and 2,440 Parasitized, while misclassifying 71 Uninfected and 51 Parasitized samples indicating strong overall classification with few errors. The sample classification using the ResNet18+TTA model is shown in Figure \ref{fig:sample_prediction}, where it correctly predicted all four samples.

\begin{figure}[H]
\begin{center}
\includegraphics[width=0.9\textwidth, trim=0cm 0cm 0cm 0cm, clip]{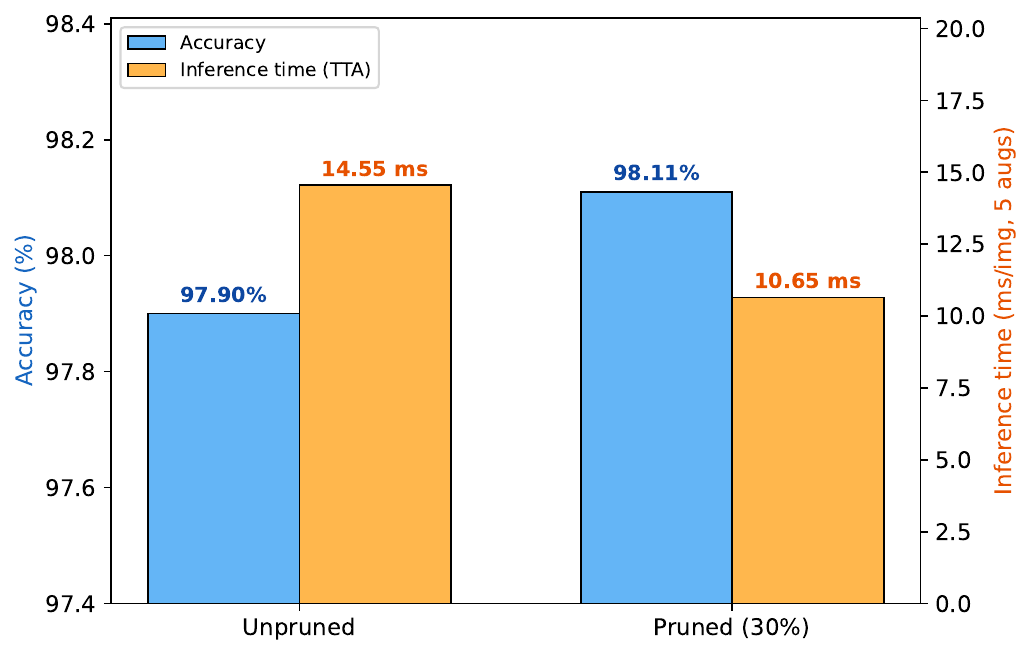}
\caption{ResNet18+TTA model under pruning performance comparison (before and after 30\% pruning) in terms of accuracy and inference speed}
\label{fig:pruned_comparision}
\end{center}
\end{figure}

To reduce the computational cost, the ResNet18+TTA model is further fine-tuned by pruning 30\% of the weights which reduces the inference time across 5 TTA augmentations from 14.55 ms to 10.65 ms and also improved accuracy by additional 0.21\% as shown in Figure \ref{fig:pruned_comparision}. The unpruned base line here (97.90\%) is within the noise floor of the test set and differs from Table \ref{tab:ablation} (97.96\%) by 0.06 pp.

The comparison of the ResNet18+TTA model with some of the state-of-the art study that uses the same dataset grouped by patient Id is shown in Table \ref{tab:benchmarking_nih}.

\begin{table}[H]
\caption{Benchmarking of studies that uses NIH malaria dataset grouped by patient id.}
\label{tab:benchmarking_nih}
\begin{center}
\small
\setlength{\tabcolsep}{6pt} % Provides clean, readable spacing for a 4-column layout
\renewcommand{\arraystretch}{1.4} 
\begin{tabular}{lllc}
\hline
\textbf{Study} & 
\textbf{Year} & 
\textbf{Model} & 
\textbf{\shortstack{Accuracy\\(in \%)}} \\
\hline
Hou et al. \cite{hou2026} & 2026 & MalariaNet & 95.6 \\
Laghari et al. \cite{laghari2025}                   & 2025 & ResNet-101 & 89.0 \\
\textbf{Our study}            & \textbf{2026} & \textbf{ResNet18+TTA} & \textbf{97.96} \\
\hline
\end{tabular}
\end{center} % 
\end{table}

\begin{figure}[h!]
\begin{center}
\includegraphics[width=0.9\textwidth, trim=0cm 0cm 0cm 0cm, clip]{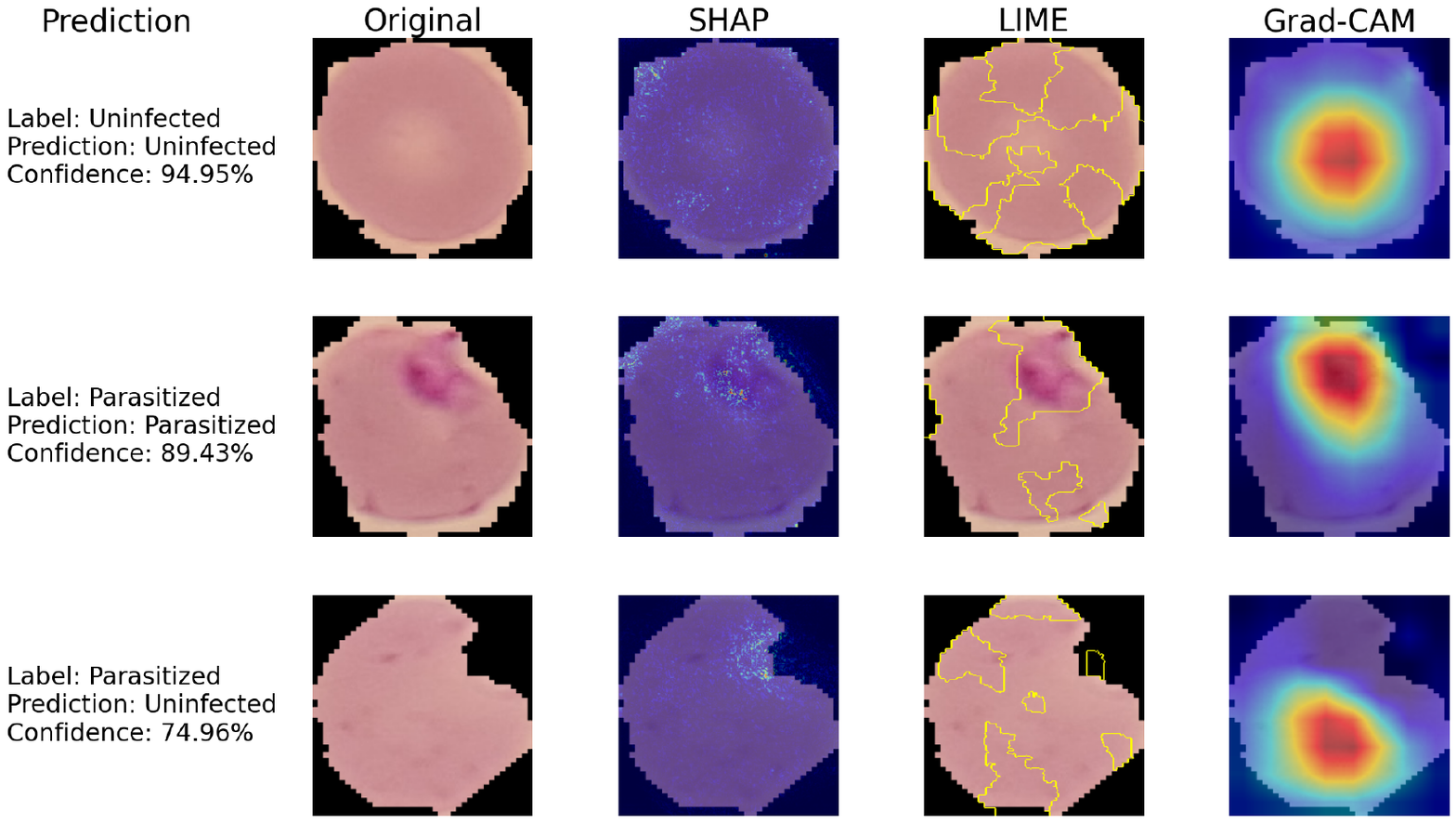}
\caption{Use of SHAP, LIME, and Grad-CAM to explain some sample detection}
\label{fig:explainable_ai}
\end{center}
\end{figure}

Our ResNet18+TTA model outperformed the state of the art study. To further assess the effectiveness of the model, we applied three complementary explainability techniques including Grad-CAM, SHAP, and LIME and visualized their output in Figure \ref{fig:explainable_ai}. We selected three representative samples: one Uninfected image followed by two Parasitized images. The model correctly classified the first two samples but misclassified the final Parasitized sample as Uninfected.

\begin{figure}[h!]
\begin{center}
\includegraphics[width=0.9\textwidth]{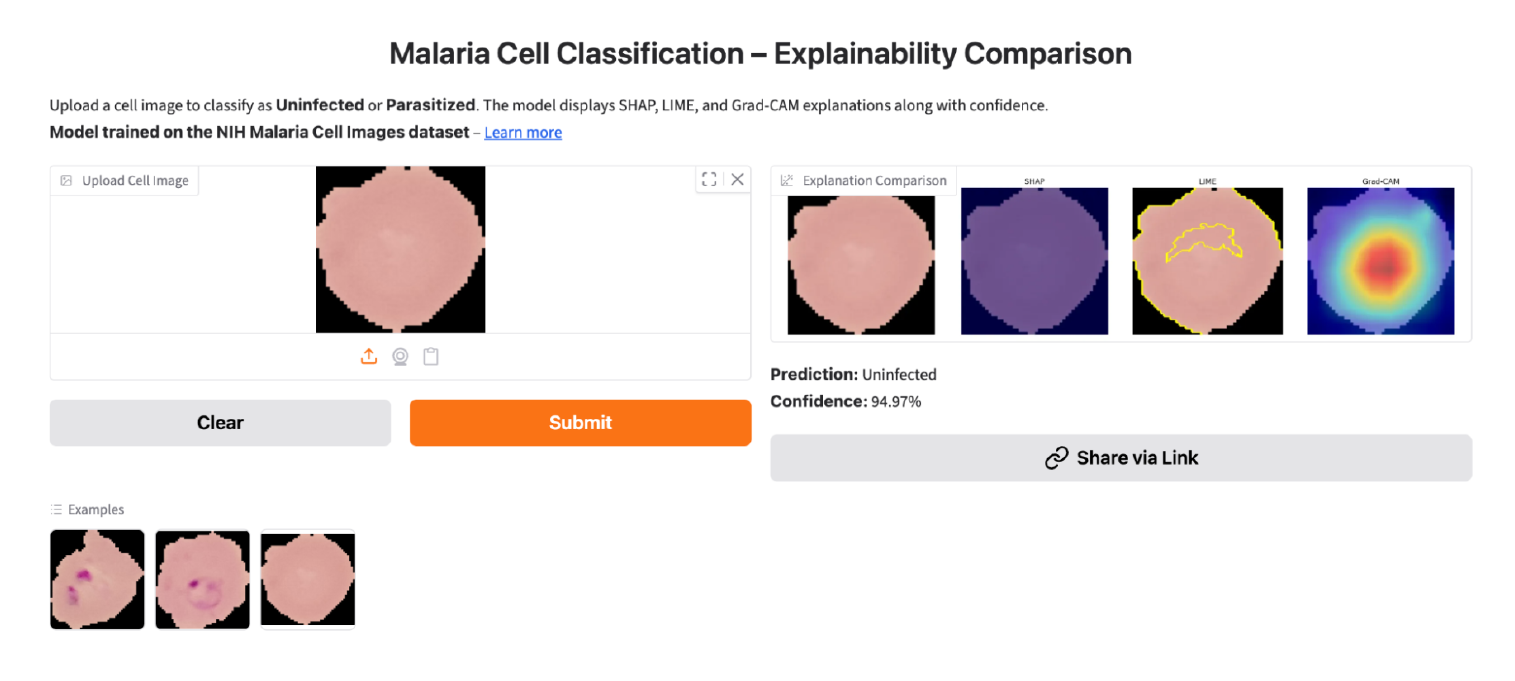}
\caption{Model deployment for real world classification of infected and uninfected malaria cells}
\label{fig:realworld_deployment}
\end{center}
\end{figure}

Across the correctly classified samples, the attribution patterns produced by SHAP, LIME, and Grad-CAM were well aligned and mutually reinforced. For the first (Uninfected) sample, SHAP displayed a diffuse, speckled attribution pattern across the cell body without a dominant focal region, LIME highlighted coherent superpixel segments indicating the prediction, and Grad-CAM emphasized a subtle region of interest where the color is different. A similar level of agreement was observed for the second (Parasitized) sample: SHAP produced  a dense, spatially coherent cluster over the diagnostically relevant region, which corresponded closely with LIME’s segmented outline and Grad-CAM’s highlighted region.
The third sample presents a more revealing case. Here, the model incorrectly predicted an Uninfected label for a Parasitized image, and the explainability outputs clearly reflect this confusion. SHAP concentrated attribution in the upper-right portion of the image, LIME highlighted multiple disparate regions, and Grad-CAM focused on the lower portion of the cell. This divergence across methods indicates a lack of coherent internal reasoning and suggests that such cases should be flagged for deeper diagnostic investigation. An web application (prototype) is built and deployed in Hugging Face Space for the general public usage as shown in Figure \ref{fig:realworld_deployment}. The clinical or user validation were not performed and its effectiveness on real-world needs to be tested. 

\section{Conclusions}
\label{Con}
In this study, we explored and benchmarked various architectures for malaria identification including ResNet18, MobileNetV2, EfficientNet-B2, ResNet101 and VGG19 using a group aware split that keeps files sharing the same identifier prefix together split to prevent the cell level leakage of the NIH malaria dataset. Furthermore, we proposed a modified ResNet18 model with custom classification header with TTA model and utilized XAI to analyze its explainability and overall performance.

From our experiments, MobileNetV2 achieved the smallest size (8.49 MB), fastest inference time (1.35 ms) and showed moderate accuracy (96.85\%). Whereas, the biggest architectures such as ResNet101 and VGG19 output larger size models 162.14 MB to 532.45 MB respectively. The EfficientNet-B2 showed 97\% accuracy with a moderate size of 29.39 MB and 2.35 ms inference speed.
Our ResNet18+TTA model showed the best accuracy (97.96\%), AUC (0.996) but suffered a high inference time (13.32 ms). Furthermore, pruning improved the accuracy of ResNet18+TTA model, reaching up to 98.11\% and reducing the inference time to 10.65 ms. These results shows the accuracy–efficiency trade-off is not monotonic with model size.

The detection of the samples were also explored using SHAP, Grad-CAM and LIME methods which provided insights on the model decision process. The methods agreed on the correctly classified cases, reinforcing confidence with the models decision. On misclassifications, they diverged highlighting model uncertainty and showing conflicting explanations can serve as a red flag for manual review. Further work needs to be done on validating these findings on different datasets and evaluate XAI faithfulness against expert annotations.

\section*{Data availability statement}
The dataset used in this study is available in \href{https://lhncbc.nlm.nih.gov/LHC-downloads/downloads.html#malaria-datasets}{NIH dataset library}, and the code used in the experiments can be available upon request.

\end{document}

%% file: elksty.tex
\def\E{\ifmmode{\mathbb E}\else{$\mathbb E$}\fi} %natural numbers
\def\N{\ifmmode{\mathbb N}\else{$\mathbb N$}\fi} %natural numbers
\def\R{\ifmmode{\mathbb R}\else{$\mathbb R$}\fi} %real numbers
\def\Q{\ifmmode{\mathbb Q}\else{$\mathbb Q$}\fi} %rational numbers
\def\C{\ifmmode{\mathbb C}\else{$\mathbb C$}\fi} %complex numbers
\def\H{\ifmmode{\mathbb H}\else{$\mathbb H$}\fi} %complex numbers
\def\Z{\ifmmode{\mathbb Z}\else{$\mathbb Z$}\fi} %integers
\def\P{\ifmmode{\mathbb P}\else{$\mathbb P$}\fi} %real numbers
\def\T{\ifmmode{\mathbb T}\else{$\mathbb T$}\fi} %real numbers
\def\SS{\ifmmode{\mathbb S}\else{$\mathbb S$}\fi} %real numbers
\def\DD{\ifmmode{\mathbb D}\else{$\mathbb D$}\fi} %real numbers

\newcommand{\bse}{\begin{subequations}}
\newcommand{\ese}{\end{subequations}}
\newcommand{\ben}{\begin{enumerate}}
\newcommand{\een}{\end{enumerate}}
\newcommand{\bens}{\begin{enumerate*}}
\newcommand{\eens}{\end{enumerate*}}
\newcommand{\be}{\begin{equation}}
\newcommand{\ee}{\end{equation}}
\newcommand{\bea}{\begin{eqnarray}}
\newcommand{\eea}{\end{eqnarray}}
\newcommand{\baa}{\begin{eqnarray*}}
\newcommand{\eaa}{\end{eqnarray*}}
\newcommand{\bc}{\begin{center}}
\newcommand{\ec}{\end{center}}

\theoremstyle{corollary}

\theoremstyle{lemma}

\theoremstyle{proposition}

\theoremstyle{axiom}

\theoremstyle{conjecture}

\theoremstyle{example}

\theoremstyle{definition}
\theoremstyle{remark}